\pdfoutput=1
\documentclass[lettersize,journal]{IEEEtran}

\usepackage{amsmath,amsfonts,amssymb}
\usepackage{algorithmic}
\usepackage{algorithm}
\usepackage{array}
\usepackage{booktabs}
\usepackage{cite}
\usepackage{graphicx}
\usepackage{textcomp}
\usepackage{stfloats}
\usepackage{flafter}
\usepackage{url}
\usepackage{makecell}
\usepackage{tikz}
\usepackage[capitalize,noabbrev]{cleveref}
\usetikzlibrary{arrows.meta,positioning,fit,calc,backgrounds}

\graphicspath{{figures/}}
\crefname{figure}{Fig.}{Figs.}
\Crefname{figure}{Fig.}{Figs.}
\crefname{table}{Table}{Tables}
\Crefname{table}{Table}{Tables}
\crefname{equation}{(}{(}
\creflabelformat{equation}{#2#1#3}

\newcommand{\Pv}{P_{\mathrm{v}}}
\newcommand{\Phat}{\hat{P}_{\mathrm{v}}}
\newcommand{\pBH}{\hat{P}_{BH}}
\newcommand{\RelErr}{\mathrm{Rel.\ Err.}}

\begin{document}

\title{CAHR-Net: Condition-Adaptive Hysteresis Reconstruction for Compact and Interpretable Magnetic Core Loss Modeling}

\author{Chunye~Gong\textsuperscript{\dag} and~Cong~Yao\textsuperscript{\dag}%
\thanks{Manuscript received Month~xx, 2026; revised Month~xx, 2026. This work was supported in part by the National Key Research and Development Program of China under Grant 2025YFB3003605 and in part by the National Natural Science Foundation of China under Grant U2530207. \emph{(\dag~Chunye Gong and Cong Yao contributed equally to this work.)} \emph{(Corresponding author: Cong Yao.)}}
\thanks{Chunye Gong and Cong Yao are with the College of Computing, National University of Defense Technology, Changsha 410073, China, also with the National Supercomputer Center in Tianjin, Tianjin 300457, China, and also with the Laboratory of Digitizing Software for Frontier Equipment, National University of Defense Technology, Changsha 410073, China (e-mail: gongchunye@nudt.edu.cn; ycong744@gmail.com).}}

\markboth{IEEE Open Journal of Power Electronics,~Vol.~XX, 2026}%
{Gong and Yao: CAHR-Net: Condition-Adaptive Hysteresis Reconstruction}

\maketitle

\begin{abstract}
Magnetic core loss originates in the hysteresis loop: the energy dissipated per excitation cycle equals the loop area, and frequency, temperature, and waveform shape set the loss by reshaping the loop geometry. Most existing models, however, let these operating conditions act only on a terminal scalar---empirical equations fold them into fitted exponents, while data-driven predictors append them to encoded features---so no intermediate hysteresis representation remains for the conditions to reshape. This paper proposes a condition-adaptive hysteresis reconstruction network (CAHR-Net) that injects the operating conditions where they physically act. The proposed method preserves the physically interpretable chain from flux density waveform to magnetic field reconstruction, hysteresis-loop area integration, and power loss estimation. Different from scalar concatenation or terminal correction, CAHR-Net uses feature-wise linear modulation to inject frequency, temperature, and waveform statistics into the intermediate reconstruction representation. A matched large-batch training protocol based on AdamW, cosine learning-rate scheduling, and a staged reconstruction-loss-to-power-loss objective is further reported, because the modulation pathway takes effect only within it. Experiments on the MagNet final A--E material protocol show that CAHR-Net attains an average $p95$ relative error of $6.89\%$ with only 1874 parameters---the lowest average $p95$ among all compared methods, together with a lower worst-material $p95$ than the strongest black-box solution at about $48\times$ fewer parameters---and reduces the average $p95$ of the physical reconstruction backbone from $7.47\%$ to $6.89\%$ and the $p95$ of material~D, the most difficult material, from $16.40\%$ to $14.87\%$. Ablation and condition-slice analyses indicate that the improvement comes from the coupling of physical loop reconstruction, structured condition modulation, and the matched optimization trajectory.
\end{abstract}


\section{Introduction}
\IEEEPARstart{M}{agnetic} core loss is not an abstract scalar response: in each excitation cycle, the energy dissipated in the core equals the area of the $B$--$H$ hysteresis loop, and this loss largely dictates the thermal design, power density, and efficiency of high-frequency converters. Operating conditions govern the loss through the geometry of this loop---frequency broadens it, temperature rescales its amplitude and local slope, and the flux-density waveform sets the trajectory along which it is traversed. Every core loss model is therefore, implicitly, an answer to the question of where the operating conditions should enter. The classical Steinmetz equation \cite{steinmetz1892} answers it by collapsing the loop into a compact power law whose fitted exponents absorb the conditions, and successive refinements have extended this answer to nonsinusoidal excitation by adding parameters to the same scalar law \cite{reinert2001_mse,venkatachalam2002_igse,muhlethaler2012_i2gse,guillod2023_cwh,arruti2024_cigse}. The collapse, however, is lossy: fixed functional forms cannot track how arbitrary waveforms, wide temperature ranges, and material-specific nonlinearities jointly deform the loop.

This difficulty is fundamental rather than parametric. Maxwell's equations and finite-element field solvers already describe the linear behavior of conductors and the geometric and thermal aspects of a component with high fidelity; what resists compact modeling is the strongly nonlinear, history-dependent magnetization that generates the loop itself, compounded by the dispersion that material composition and manufacturing introduce at the component level. Hysteresis formalisms such as the Preisach model \cite{preisach1935} and the Jiles--Atherton model \cite{jiles1986} keep the loop as the modeling object, but their semi-empirical parameters are difficult to identify consistently across wide frequency, temperature, and waveform ranges. Core loss modeling has therefore long faced a dilemma: preserve the loop and struggle to fit it, or collapse the loop and lose the very geometry through which the operating conditions act.

Public datasets and modeling challenges \cite{magnet2023,magnet_ai2023,why_magnet2023,magnet_challenge2025} have recently made a third route practical: learning the waveform-to-loss map directly from large-scale measurements. Neural predictors built on waveform encoding, multimodal fusion, and physics-inspired structures \cite{pi_mff_cn2024,empinn2025,mminn2024,xiao2026_pinn}, together with knowledge-aware networks \cite{deng2024_kann}, history-dependent hysteresis operators \cite{huang2025_hdpi}, adversarial generation \cite{shen2024_cgan}, and cross-material transfer \cite{chen2025_mkmmd}, have pushed accuracy well beyond the empirical family. Yet most of these models achieve accuracy by abandoning the loop altogether, and they inherit a subtler form of the same misplacement: frequency, temperature, and flux-density statistics are appended to a feature vector or concatenated at the regression head, so the model can observe which condition produced a sample, while no intermediate hysteresis representation remains for the conditions to reshape.

Gray-box reconstruction restores the missing object. The HARDCORE framework \cite{hardcore2025} first reconstructs the magnetic field waveform and then integrates the $BH$ loop area, showing that keeping the loop inside the model yields a favorable accuracy--complexity tradeoff under arbitrary waveforms. Even here, however, the operating conditions still enter as appended scalars: the object is right, but the conditioning pathway is not. We observe that the leading effects of frequency and temperature on the loop---broadening, tilting, and amplitude scaling---are naturally expressed as a scaling and shifting of the representation from which the loop is generated. This suggests encoding the operating conditions into feature-wise scale and shift parameters that act directly on the intermediate reconstruction representation. CAHR-Net, the condition-adaptive hysteresis reconstruction network proposed in this paper, realizes this principle: a dilated temporal-convolutional encoder \cite{tcn2018} extracts the waveform skeleton, feature-wise linear modulation (FiLM) \cite{film2018} lets the conditions reshape the hidden representation before the magnetic field waveform is reconstructed, and a staged reconstruction-to-loss objective with matched large-batch optimization allows the modulation branch to leave the near-identity regime and become effective.

The contributions of this paper follow from this single design question---where should the operating conditions enter the model---and are summarized as follows.
\begin{itemize}
\item \emph{The object: a physically interpretable reconstruction backbone.} A lightweight architecture keeps the hysteresis loop as the modeling object for arbitrary-waveform core loss estimation, following the interpretable path $B(t)\rightarrow \hat{H}(t)\rightarrow BH\rightarrow \Phat$ rather than direct terminal regression.
\item \emph{The pathway: condition-adaptive FiLM modulation.} Frequency, temperature, and waveform statistics are encoded into feature-wise scaling and shifting parameters that act directly on the intermediate hysteresis reconstruction representation. This pathway admits an electromagnetic reading---scaling matches loop broadening and amplitude rescaling, shifting matches bias and remanence drift---that scalar concatenation and terminal correction do not provide.
\item \emph{The evidence: a favorable accuracy--parameter--interpretability tradeoff.} Under the unified MagNet final A--E protocol, CAHR-Net attains an average $p95$ relative error of $6.89\%$ with only 1874 parameters, the lowest average $p95$ among all compared methods---and a lower worst-material $p95$ than the strongest black-box entry at about $48\times$ fewer parameters---and it reduces the average $p95$ of the physical reconstruction backbone from $7.47\%$ to $6.89\%$ and the $p95$ of material~D, the most difficult material, from $16.40\%$ to $14.87\%$.
\end{itemize}

The rest of this paper is organized as follows. \Cref{sec:method} develops CAHR-Net, from the problem formulation to the loop reconstruction chain, the condition modulation pathway, and the deployment-oriented inference optimization. \Cref{sec:setup} describes the experimental setup, including the network configuration and training protocol, \Cref{sec:results} presents the results and analyses, and \Cref{sec:conclusion} concludes the paper.

\section{Methodology}
\label{sec:method}

\subsection{Problem Formulation and Method Overview}
Given a single-period flux density waveform
\begin{equation}
  \mathbf{B}=\{B_t\}_{t=1}^{N}, \quad N=1024,
\end{equation}
and a scalar operating vector $\mathbf{s}$ containing frequency, temperature, and waveform-level statistics, the goal is to predict the volumetric core loss $\Pv$. A direct map
\begin{equation}
  \Phat=f_{\theta}(\mathbf{B},\mathbf{s})
\end{equation}
treats the loss as an isolated scalar output and leaves no intermediate representation for the operating conditions to act on. CAHR-Net instead uses a physically structured intermediate target that keeps such a pathway open: it reconstructs the magnetic field waveform $\hat{H}(t)$, computes a loop-area-based initial estimate, and then applies a lightweight residual correction in the logarithmic loss domain.

\begin{figure*}[!tbp]
\centering
\includegraphics[width=0.72\textwidth]{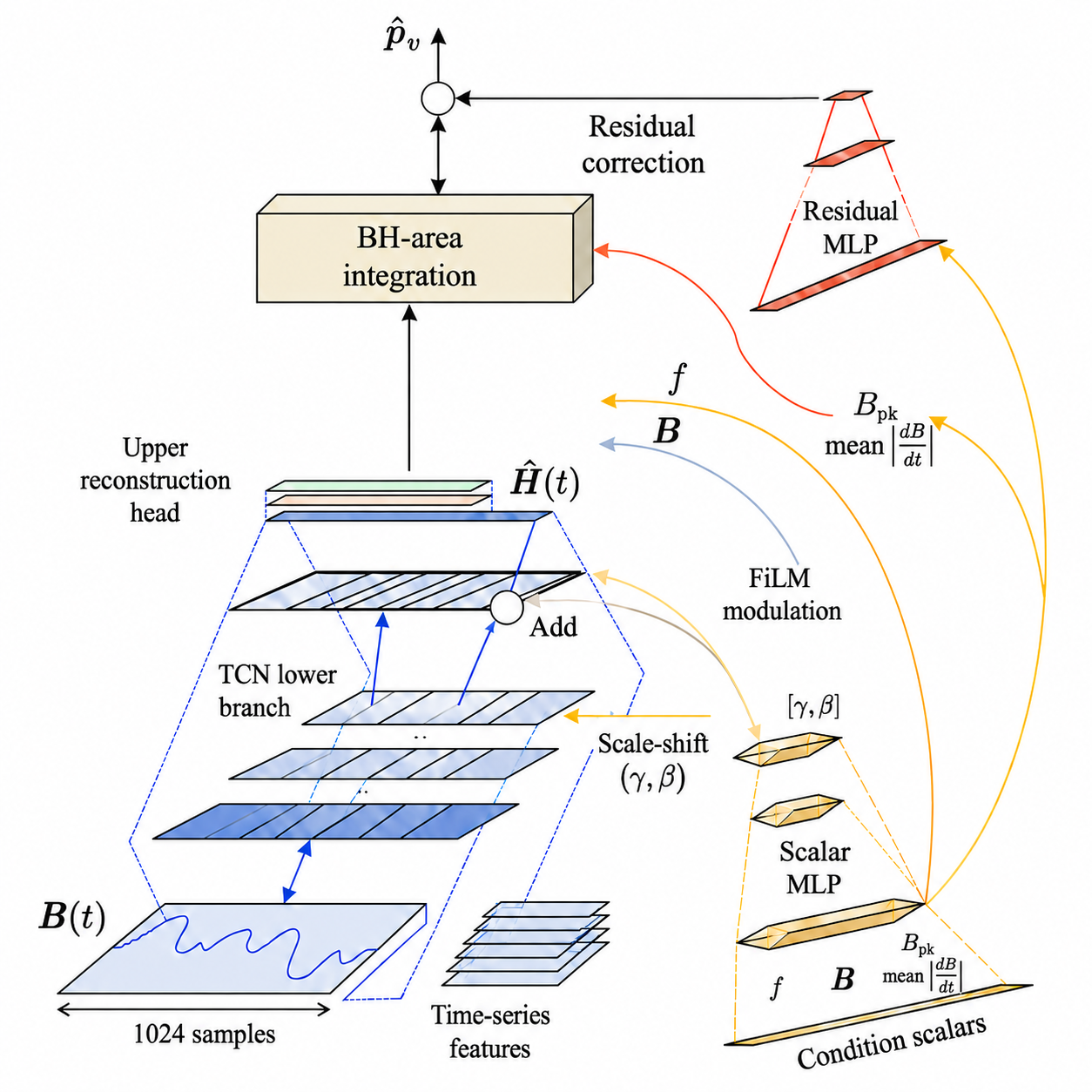}
\caption{Overall architecture of CAHR-Net. The operating vector $\mathbf{s}$ is encoded into the FiLM pair $[\gamma,\beta]$, which rescales ($\odot$, by $1+\alpha\gamma$) and then shifts ($+$, by $\beta$) eleven of the twelve encoder channels before the magnetic-field reconstruction head; one raw channel bypasses modulation. The reconstructed field $\hat{H}(t)$ drives the $BH$-area integration, and a residual multilayer perceptron corrects the loop-area estimate in the logarithmic loss domain to produce $\hat{P}_{\mathrm{v}}$.}
\label{fig:chapter6-cahr-arch}
\end{figure*}

\Cref{fig:chapter6-cahr-arch} shows the resulting architecture. CAHR-Net contains a temporal convolutional branch for waveform encoding, a scalar multilayer perceptron for condition encoding, a FiLM block \cite{film2018} that couples the two, a magnetic-field reconstruction head with $BH$ area integration, and a residual power-loss correction head.

\subsection{Physical Loop Reconstruction Chain}
\label{sec:loop-chain}
The temporal encoder first extracts a waveform skeleton from the normalized 1024-point flux density sequence,
\begin{equation}
  \mathbf{F}_0=\mathcal{E}_{B}\!\left(B(t)\right),
\end{equation}
where $\mathcal{E}_{B}(\cdot)$ denotes a stack of dilated temporal convolution blocks \cite{tcn2018}. The scalar branch maps the operating vector to modulation parameters. After modulation, the reconstruction head generates
\begin{equation}
  \hat{H}(t)=\mathcal{G}\!\left(B(t),\mathbf{s};\gamma,\beta\right).
\end{equation}

The reconstructed field waveform is converted to an initial power-loss estimate through discrete loop-area integration,
\begin{equation}
  \pBH=f\,b_{\lim}h_{\lim}\operatorname{trapz}\!\left(\hat{H},B\right),
  \label{eq:cahr-bh}
\end{equation}
where $f$ is the excitation frequency, $b_{\lim}$ and $h_{\lim}$ are the scale factors used to restore physical magnitudes, and $\operatorname{trapz}(\cdot)$ denotes trapezoidal integration.

The final prediction is obtained in the logarithmic domain:
\begin{equation}
  \hat{y}_{P}=\log\!\left(\pBH\right)+
  \Delta\!\left([\mathbf{s},\log(\pBH)]\right),
  \label{eq:cahr-logp}
\end{equation}
\begin{equation}
  \Phat=\exp(\hat{y}_{P}).
  \label{eq:cahr-ploss}
\end{equation}
In implementation, the residual head uses a standardized form of $\log(\pBH)$ for numerical stability. \Cref{eq:cahr-bh,eq:cahr-logp,eq:cahr-ploss} keep the prediction chain anchored to the physical relation between the hysteresis loop and core loss while allowing a small learned correction for scale and integration bias.

Both the reconstruction target and the scale factors in \Cref{eq:cahr-bh} are anchored in measured data. For every record, the MagNet database provides the measured magnetic field waveform $H(t)$, obtained from the sensed excitation current through Amp\`ere's law, alongside the flux density waveform obtained from the induced voltage \cite{magnet2023}. The reconstruction loss defined in \Cref{sec:training} therefore supervises $\hat{H}(t)$ with measured waveforms rather than with model-generated references, following the same protocol as HARDCORE \cite{hardcore2025}. The factor $b_{\lim}$ is the maximum absolute flux density over the training records of the material, and $h_{\lim}$ is the corresponding field normalization bound, capped at $150~\mathrm{A/m}$ and rescaled per record by the record's relative flux amplitude, so that \Cref{eq:cahr-bh} restores physical magnitudes from normalized waveforms.

\subsection{Condition-Adaptive FiLM Modulation}
\label{sec:film}
The key difference between CAHR-Net and scalar-concatenation models is the location and form of condition injection. Let $\mathcal{E}_{s}(\cdot)$ be the scalar encoder. The modulation parameters are generated by
\begin{equation}
  [\gamma,\beta]=\Psi\!\left(\mathcal{E}_{s}(\mathbf{s})\right),
\end{equation}
where $\Psi(\cdot)$ projects the condition embedding into the feature modulation space. For an intermediate representation $\mathbf{F}$, CAHR-Net applies
\begin{equation}
  \widetilde{\mathbf{F}}=(1+\alpha\gamma)\odot\mathbf{F}+\beta,
  \label{eq:cahr-film}
\end{equation}
where $\odot$ is channel-wise multiplication and $\alpha$ controls modulation strength. The default setting uses $\alpha=0.1$.

The form of \Cref{eq:cahr-film} mirrors the two dominant ways in which operating conditions deform the measured hysteresis loop. As frequency rises, eddy-current and excess contributions widen the loop and rescale its local slope, which is a multiplicative deformation of the trajectory amplitude; temperature shifts the permeability and coercivity operating point, which acts closer to a translation of the loop. The FiLM pair reproduces these two degrees of freedom on the hidden representation: $\gamma$ applies channel-wise rescaling, matching loop broadening and amplitude scaling, while $\beta$ applies channel-wise shifting, matching bias and remanence drift. The same waveform pattern can therefore be amplified, suppressed, or translated under different operating conditions before $\hat{H}(t)$ is reconstructed. Additive bias injection, as used in the backbone, covers only the translational degree of freedom, and input-level concatenation provides neither in a structured form. We do not claim a one-to-one identification between individual FiLM parameters and specific loop features; the correspondence holds at the level of transformation families, and it is consistent with the ablation in \Cref{fig:chapter6-cahr-ablation-pareto}(a), where replacing FiLM with bias-only injection removes most of the gain under the matched optimization protocol.

\subsection{Deployment-Oriented Inference Optimization}
\label{sec:deployment}
The 1874-parameter budget suggests that CAHR-Net is inexpensive to deploy, but parameter count alone does not determine the latency realized in practice. In the intended deployment scenarios---loss evaluation inside iterative converter design loops, large design-space sweeps, and online estimation on general-purpose processors without an accelerator---the model is queried one operating point at a time on a CPU, and for a network of this size the per-query cost is dominated by framework dispatch rather than by arithmetic. CAHR-Net therefore admits an accuracy-neutral inference optimization that lowers this per-query cost by changing only the execution backend, without modifying weights, architecture, or numerical precision.

The optimization exports the trained model to the ONNX format and serves it with ONNX Runtime~1.18.1 \cite{onnxruntime2021}. The export is not a push-button step: the loop-area integration of \Cref{eq:cahr-bh} relies on a trapezoidal-integration operator that has no ONNX equivalent. A thin export wrapper therefore rewrites the integration as an explicit trapezoidal sum built from elementary slicing, addition, multiplication, and reduction operations, and bakes the per-material normalization constants into the graph, so that the exported graph is self-contained. Two numerical gates guard the export. First, the wrapper is verified to be bit-identical to the original TorchScript model, with a maximum absolute output difference of exactly zero. Second, the exported model under ONNX Runtime is compared against the original over all 7651 material-A test samples: the maximum deviation is $|\Delta \log \Phat| < 8\times 10^{-6}$, and the test-set $p95$ relative error remains $5.86\%$, unchanged to two decimals. The optimization is thus accuracy-neutral both by construction and by measurement; its latency benefit is quantified in \Cref{sec:latency-eval}.

\section{Experimental Setup}
\label{sec:setup}
\subsection{Dataset, Protocol, and Metrics}
Experiments use the MagNet final-stage A--E material protocol \cite{magnet2023,magnet_challenge2025}. Each material is trained and tested independently on its corresponding final split. The input waveform contains 1024 samples per period, and the scalar vector contains frequency, temperature, and waveform statistics as specified in \Cref{sec:config}.

Prediction quality is measured by the relative error
\begin{equation}
  \RelErr=\left|\frac{\Phat-\Pv}{\Pv}\right|\times 100\%,
\end{equation}
and reported as the average relative error, the average $p95$ and $p99$ percentiles across the five materials, and the worst-material $p95$ \cite{tofallis2015}. In all experiments reported below, the worst-material $p95$ occurs on material~D, so ``worst $p95$'' and ``material-D $p95$'' denote the same quantity. The $p95$-oriented view is adopted because thermal design is usually more sensitive to high-risk tail errors than to mean behavior; an average $p95$ relative error below $10\%$ is generally regarded as strong under the MagNet evaluation protocol \cite{magnet_challenge2025}. All reported results follow the single-model reporting convention of the MagNet Challenge and HARDCORE \cite{magnet_challenge2025,hardcore2025}, using one trained model per method under the released protocol.

\subsection{Compared Methods}
Three groups of baselines are considered. First, empirical models, including SE, MSE, GSE, and iGSE, are re-fitted under the same A--E protocol. SE is fitted by logarithmic least squares, whereas MSE, GSE, and iGSE use bounded nonlinear least squares with five multi-start restarts. Second, representative public challenge methods are summarized from the MagNet Challenge report \cite{magnet_challenge2025}. Third, internal models are evaluated under the same experimental chain, including the HARDCORE physical reconstruction backbone~\cite{hardcore2025}, the same backbone trained with AdamW and cosine scheduling (HARDCORE + AWC), SE-TCN, FiLM-TCN, CAHR-Net, and width or combination variants.

\subsection{Network Configuration}
\label{sec:config}
\Cref{tab:cahr-config} lists the complete layer-by-layer configuration of CAHR-Net; the parameter subtotals add up to the 1874 parameters reported throughout the paper. The waveform encoder receives five input channels---the globally normalized flux density waveform, a per-record normalized copy, its first and second time derivatives, and a saturation-emphasizing tangent transform \cite{hardcore2025}. The operating vector $\mathbf{s}\in\mathbb{R}^{11}$ collects the logarithmic frequency, the temperature, four waveform-shape indicator variables, the peak-to-peak flux density and the mean absolute flux slew rate together with their logarithms, and the fundamental period. All convolutions use circular padding, consistent with the periodic single-period input, and the three temporal blocks use kernel size 9 with dilation rates 4, 8, and 16. FiLM acts at the interface between the waveform encoder and the reconstruction head: eleven of the twelve encoder channels are modulated by $\gamma,\beta\in\mathbb{R}^{11}$, while the first channel passes through unmodulated so that a raw waveform pathway is always preserved.

\begin{table}[!tb]
\caption{Layer-by-Layer Configuration of CAHR-Net}
\label{tab:cahr-config}
\centering
\small
\setlength{\tabcolsep}{3pt}
\renewcommand{\arraystretch}{1.15}
\begin{tabular}{@{}l l r@{}}
\toprule
Stage & Configuration & Params \\
\midrule
Waveform encoder $\mathcal{E}_{B}$ & Conv1d $5{\rightarrow}12$, $k{=}9$, $d{=}4$, tanh & 552 \\
Condition encoder $\mathcal{E}_{s},\Psi$ & Linear $11{\rightarrow}22$, tanh, $[\gamma,\beta]$ & 264 \\
FiLM modulation & \Cref{eq:cahr-film}, 11 of 12 channels & 0 \\
Reconstruction head $\mathcal{G}$ & Conv1d $12{\rightarrow}8$, $k{=}9$, $d{=}8$, tanh & 872 \\
 & Conv1d $8{\rightarrow}1$, $k{=}9$, $d{=}16$ & 73 \\
$BH$ integration & Trapezoidal, \Cref{eq:cahr-bh} & 0 \\
Residual head $\Delta$ & Linear $12{\rightarrow}8$, tanh; Linear $8{\rightarrow}1$ & 113 \\
\midrule
Total & & 1874 \\
\bottomrule
\end{tabular}
\end{table}

CAHR-Net is fully specified by \Cref{tab:cahr-config} and the training protocol below; HARDCORE is included in \Cref{tab:internal-main} only as a reconstruction baseline under the same A--E evaluation protocol.

\subsection{Training Protocol}
\label{sec:training}
Introducing FiLM alone does not guarantee a stable gain under large-batch training. CAHR-Net therefore uses a matched optimization protocol. The magnetic-field reconstruction loss and logarithmic power-loss loss are
\begin{equation}
  \mathcal{L}_{H}=\mathrm{MSE}\!\left(\hat{H},H\right), \quad
  \mathcal{L}_{P}=\mathrm{MSE}\!\left(\hat{y}_{P},\log \Pv\right).
\end{equation}
At epoch $e$ of $E$ total epochs, the total loss is
\begin{equation}
  \mathcal{L}(e)=\left(1-\frac{e}{E}\right)\mathcal{L}_{H}
  +\frac{e}{E}\mathcal{L}_{P}.
  \label{eq:cahr-loss}
\end{equation}
This schedule prioritizes stable loop reconstruction at the beginning and gradually shifts the emphasis toward loss prediction. The ordering matters for interpretability: if the power-loss objective dominated from the start, the residual head could compensate for a physically meaningless loop, and the reconstruction chain would lose its electromagnetic reading.

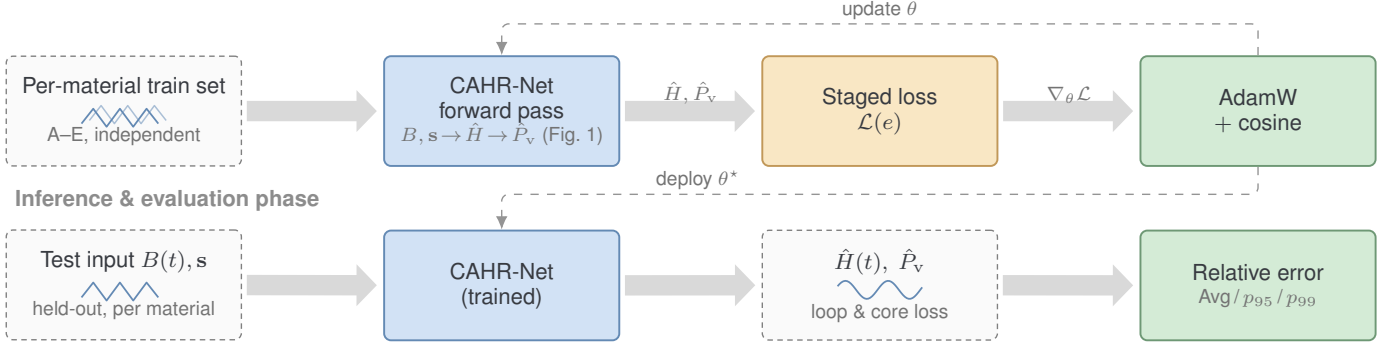
\begin{figure*}[!t]
\centering
\definecolor{waveblue}{RGB}{210,226,245}
\definecolor{condgold}{RGB}{249,231,196}
\definecolor{physgreen}{RGB}{212,235,214}
\definecolor{waveblueS}{RGB}{100,138,180}
\definecolor{condgoldS}{RGB}{193,152,89}
\definecolor{physgreenS}{RGB}{110,152,117}
\definecolor{figink}{RGB}{55,58,64}
\resizebox{\textwidth}{!}{%
\begin{tikzpicture}[
  font=\sffamily\footnotesize,
  text=figink,
  blk/.style 2 args={rounded corners=2.5pt, draw=#1, fill=#2, line width=0.7pt,
      minimum height=14mm, minimum width=30mm, align=center, inner sep=3pt},
  io/.style={rounded corners=2.5pt, draw=black!50, dash pattern=on 2pt off 1.6pt,
      fill=black!2, line width=0.5pt,
      minimum height=14mm, minimum width=30mm, align=center, inner sep=3pt},
  lbl/.style={font=\sffamily\scriptsize, text=black!55, inner sep=1.5pt},
  fat/.style={-{Triangle[length=2.6mm, width=5.2mm]}, line width=2.6mm, black!15,
      shorten >=0.8mm, shorten <=0.6mm},
  feedback/.style={->, >={Latex[length=1.7mm, width=1.5mm]},
      black!42, line width=0.5pt, dashed, rounded corners=2.5pt},
  phase/.style={font=\sffamily\bfseries\footnotesize, text=black!45},
]
\node[io] (tdata) {Per-material train set\\[8.5pt]{\scriptsize\textcolor{black!55}{A--E, independent}}};
\node[blk={waveblueS}{waveblue}, right=18mm of tdata] (fwd)
     {CAHR-Net\\forward pass\\[-0.5pt]{\scriptsize\textcolor{black!55}{$B,\mathbf{s}\!\to\!\hat H\!\to\!\hat P_{\mathrm v}$ (Fig.~\ref{fig:chapter6-cahr-arch})}}};
\node[blk={condgoldS}{condgold}, right=18mm of fwd] (sched) {Staged loss\\$\mathcal{L}(e)$};
\node[blk={physgreenS}{physgreen}, right=18mm of sched] (opt) {AdamW\\$+$ cosine};
\draw[fat] (tdata) -- (fwd);
\draw[fat] (fwd) -- (sched);
\draw[fat] (sched) -- (opt);
\node[lbl, above=0.4mm of {$(fwd.east)!0.5!(sched.west)$}] {$\hat H,\hat P_{\mathrm v}$};
\node[lbl, above=0.4mm of {$(sched.east)!0.5!(opt.west)$}] {$\nabla_{\theta}\mathcal{L}$};
\begin{scope}[shift={($(tdata.center)+(0,-0.6mm)$)}, x=1mm, y=1mm]
  \draw[waveblueS!55, line width=0.55pt, line join=round]
    (-4.4,-0.6) -- (-2.9,1.3) -- (-1.2,-1.0) -- (0.5,1.3) -- (2.2,-1.0) -- (3.9,1.3) -- (5.4,-0.6);
  \draw[waveblueS, line width=0.65pt, line join=round]
    (-5.4,-1.0) -- (-3.9,0.9) -- (-2.2,-1.4) -- (-0.5,0.9) -- (1.2,-1.4) -- (2.9,0.9) -- (4.4,-1.0);
\end{scope}
\coordinate (fby) at ($(fwd.north)+(0,4mm)$);
\draw[feedback]
  (opt.north) -- (opt.north |- fby) -- node[lbl,above]{update $\theta$} (fwd.north |- fby) -- (fwd.north);
\node[phase, anchor=south west] at ($(tdata.west |- fby)+(0,2.5mm)$) {Training phase};
\node[io, below=8mm of tdata] (tin) {Test input $B(t),\mathbf{s}$\\[8.5pt]{\scriptsize\textcolor{black!55}{held-out, per material}}};
\node[blk={waveblueS}{waveblue}, below=8mm of fwd] (infer) {CAHR-Net\\(trained)};
\node[io, below=8mm of sched] (pred) {$\hat H(t),\ \hat P_{\mathrm v}$\\[8.5pt]{\scriptsize\textcolor{black!55}{loop \& core loss}}};
\node[blk={physgreenS}{physgreen}, below=8mm of opt] (eval)
     {Relative error\\[-0.5pt]{\scriptsize\textcolor{black!55}{Avg\,/\,$p_{95}$\,/\,$p_{99}$}}};
\draw[fat] (tin) -- (infer);
\draw[fat] (infer) -- (pred);
\draw[fat] (pred) -- (eval);
\begin{scope}[shift={($(tin.center)+(0,-0.6mm)$)}, x=1mm, y=1mm]
  \draw[waveblueS, line width=0.65pt, line join=round]
    (-5.4,-1.0) -- (-3.9,0.9) -- (-2.2,-1.4) -- (-0.5,0.9) -- (1.2,-1.4) -- (2.9,0.9) -- (4.4,-1.0);
\end{scope}
\begin{scope}[shift={($(pred.center)+(0,-0.6mm)$)}, x=1mm, y=1mm]
  \draw[waveblueS, line width=0.65pt]
    plot[domain=-5.4:5.4, samples=45] (\x, {1.15*sin(\x*66+20)});
\end{scope}
\coordinate (corr) at ($(opt.south)+(0,-3.5mm)$);
\draw[feedback]
  (opt.south) -- (corr) -- node[lbl,above,pos=0.74]{deploy $\theta^{\star}$} (infer.north |- corr) -- (infer.north);
\node[phase, anchor=south west] at ($(tin.west |- tin.north)+(0,1.2mm)$) {Inference \& evaluation phase};
\end{tikzpicture}}
\caption{Training-to-inference protocol of CAHR-Net. Each material A--E is trained independently. During training, the forward pass of \Cref{fig:chapter6-cahr-arch} produces $\hat{H}$ and $\hat{P}_{\mathrm{v}}$, which are scored by the staged loss $\mathcal{L}(e)$ of \Cref{eq:cahr-loss} and optimized with AdamW and cosine scheduling; the curriculum schedule shifts the emphasis from the magnetic-field reconstruction loss $\mathcal{L}_H$ early in training to the logarithmic core-loss objective $\mathcal{L}_P$ late in training. The trained parameters $\theta^{\star}$ are then deployed on held-out inputs, and the average, $p_{95}$, and $p_{99}$ relative-error metrics are computed per material.}
\label{fig:cahr-pipeline}
\end{figure*}

The optimizer is AdamW \cite{adamw2019} with cosine learning-rate scheduling \cite{sgdr2017}, learning rate $2\times 10^{-3}$, weight decay $10^{-4}$, batch size 512, and 10000 epochs; the modulation strength $\alpha=0.1$ in \Cref{eq:cahr-film} keeps the FiLM branch close to identity at initialization. These settings are stated in full because of an empirical finding documented in \Cref{fig:chapter6-cahr-ablation-pareto}(a): under the original NAdam optimizer with step decay, FiLM behaves close to bias injection, whereas AdamW with cosine scheduling and the enlarged learning rate allows the modulation branch to leave the near-identity regime and take effect. We regard this coupling between the conditioning structure and the optimization trajectory as an empirical finding rather than a standalone contribution. \Cref{fig:cahr-pipeline} summarizes the resulting training-to-inference protocol.

\section{Results and Discussion}
\label{sec:results}
\subsection{Comparison: Empirical Equations, Challenge Entries, and the Accuracy--Parameter Frontier}
\begin{figure}[!tb]
\centering
\includegraphics[width=\linewidth]{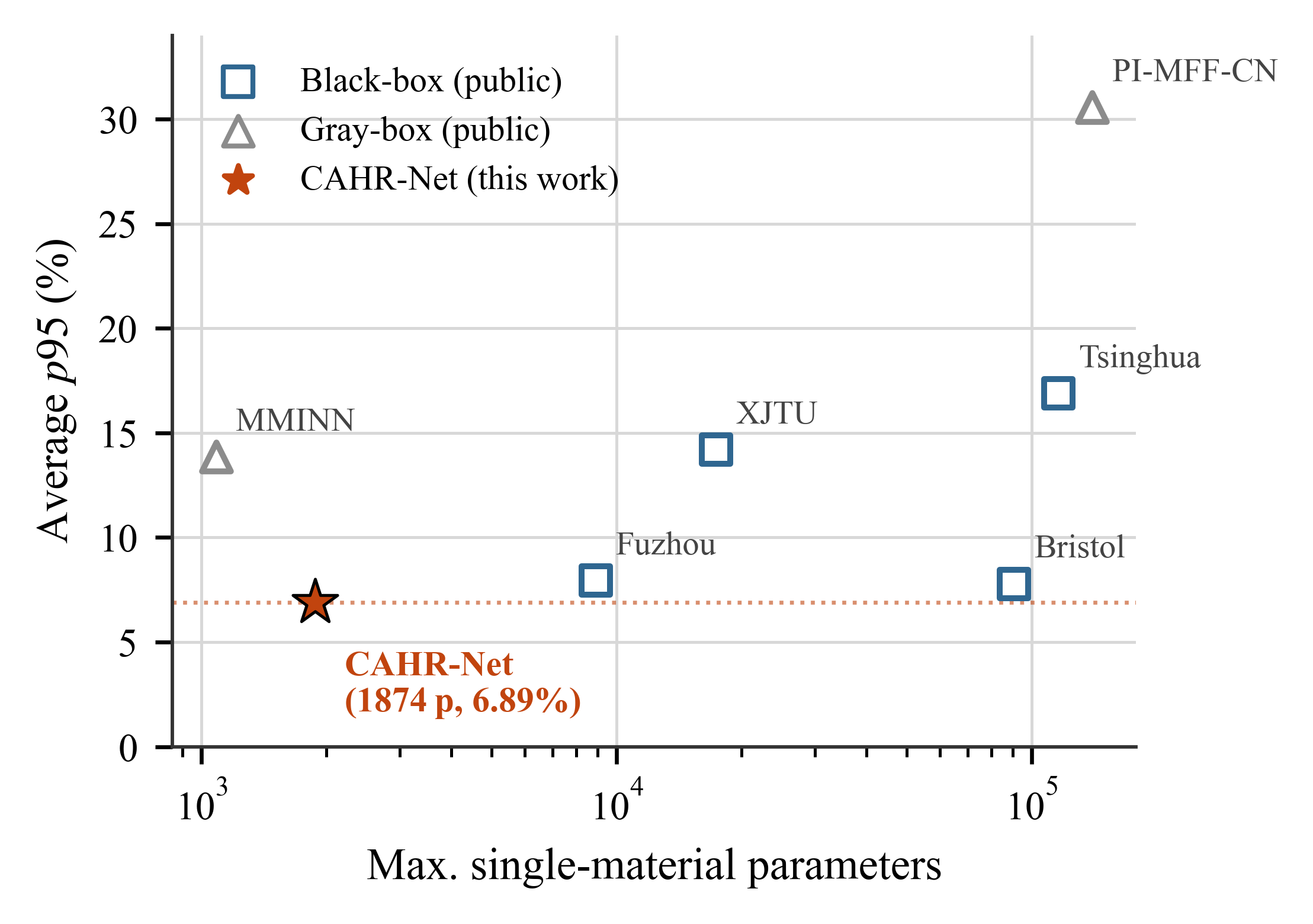}
\caption{Parameter-count and average-$p95$ relation between public representative methods and CAHR-Net.}
\label{fig:chapter6-cahr-public-scatter}
\end{figure}

\Cref{tab:unified-comparison} consolidates the external comparison into three views. \emph{Empirical equations.} Even when SE, MSE, GSE, and iGSE are fitted independently for each material, they remain in the $56.84\%$--$78.30\%$ average-$p95$ band; their limitation is therefore structural rather than a matter of parameter fitting, because a terminal scalar law cannot represent arbitrary-waveform, wide-condition effects. \emph{MagNet Challenge entries.} The public field spans a wide range, from the strongest black-box solution, Bristol, at $7.78\%$ average $p95$ with $90{,}653$ parameters, to gray-box entries that stay above $13\%$; these numbers were obtained under different implementations and possibly different evaluation protocols, so they serve as an external reference rather than a controlled ranking. \emph{Accuracy--parameter frontier.} Isolating the four most competitive operating points shows where the proposed method leads: Bristol, which reports the best accuracy yet placed third officially; Fuzhou, the official runner-up; HARDCORE, the method of the official winner Paderborn, evaluated here as the reconstruction backbone; and CAHR-Net. Note that the official final ranking weighted model size jointly with accuracy~\cite{magnet_challenge2025}, which is why the 8914-parameter Fuzhou entry outranked Bristol despite Bristol's lower error. CAHR-Net attains the lowest average $p95$ in the whole table, $6.89\%$ against Bristol's $7.78\%$ and Fuzhou's $7.94\%$, together with the lowest worst-material $p95$, $14.87\%$ against Bristol's $15.90\%$, yet reaches this with only 1874 parameters---about $1/48$ of the Bristol budget and $1/4.8$ of Fuzhou---while the gray-box models near its size stay above $13\%$ average $p95$. CAHR-Net thus occupies a region of the accuracy--parameter plane that neither the black-box nor the existing gray-box group covers, while keeping the prediction chain physically readable. \Cref{fig:chapter6-cahr-public-scatter} visualizes this relationship.

\begin{table*}[!tbp]
\caption{Unified Comparison: Empirical Equations, MagNet Challenge Entries, and the Accuracy--Parameter Frontier}
\label{tab:unified-comparison}
\centering
\small
\setlength{\tabcolsep}{4pt}
\renewcommand{\arraystretch}{1.2}
\resizebox{0.98\textwidth}{!}{%
\begin{tabular}{p{3.6cm} p{4.2cm} c c c}
\toprule
Method & Core Expression / Category & Params & Avg $p95$ (\%) & Worst $p95$ (\%) \\
\midrule
\multicolumn{5}{@{}l}{\textit{(a) Empirical equations (per-material fit, unified A--E protocol)}}\\
SE   & $k f^{\alpha} B_{\mathrm{pk}}^{\beta}$ & --- & 78.30 & 85.84 \\
MSE  & $k f^{\alpha} B_{\mathrm{pk}}^{\beta} f_{\mathrm{eq}}^{\alpha-1}$ & --- & 56.84 & 62.31 \\
GSE  & $k f^{\alpha}\Delta B^{\beta-\alpha}\int |\dot{B}|^{\alpha}$ & --- & 57.66 & 64.52 \\
iGSE & $k f^{\alpha}\sum_i \Delta B_i^{\beta}\Delta t_i^{1-\alpha}$ & --- & 61.23 & 72.42 \\
\midrule
\multicolumn{5}{@{}l}{\textit{(b) MagNet Challenge entries (as reported~\cite{magnet_challenge2025})}}\\
Bristol   & Black-box & 90653  & 7.78  & \textbf{15.90} \\
Fuzhou    & Black-box & 8914   & 7.94  & 20.70 \\
Tsinghua  & Black-box & 116061 & 16.88 & 29.90 \\
XJTU      & Black-box & 17342  & 14.20 & 30.00 \\
MMINN~\cite{mminn2024}         & Gray-box & \textbf{1084} & 13.86 & 30.70 \\
PI-MFF-CN~\cite{pi_mff_cn2024} & Gray-box & 139938 & 30.56 & 79.10 \\
\midrule
\multicolumn{5}{@{}l}{\textit{(c) Accuracy--parameter frontier}}\\
Bristol \small(best reported accuracy) & Black-box & 90653 ($48\times$) & 7.78 (+12.9\%) & 15.90 (+6.9\%) \\
Fuzhou \small(official 2nd place) & Black-box & 8914 ($4.8\times$) & 7.94 (+15.2\%) & 20.70 (+39.2\%) \\
HARDCORE \small(official 1st place; backbone) & Gray-box & 1742 ($0.93\times$) & 7.47 (+8.4\%) & 16.40 (+10.3\%) \\
\textbf{CAHR-Net} & Gray-box & 1874 & \textbf{6.89} & \textbf{14.87} \\
\bottomrule
\end{tabular}}
\par\vspace{2pt}
{\footnotesize\raggedleft Blocks (a) and (c, backbone/CAHR-Net rows) are evaluated under the unified protocol of this work~(\Cref{sec:setup}); block (b) and the Bristol/Fuzhou rows of (c) are as reported in the original sources~\cite{magnet_challenge2025}, so the cross-group comparison is an external reference rather than a controlled ranking. Official placements refer to the challenge's final ranking, which jointly weighted accuracy and model size. Parenthesized values in block (c) are relative to CAHR-Net: the parameter count as a multiple of 1874, and the two $p95$ columns as the relative increase over $6.89\%$ and $14.87\%$, respectively. The Bristol entry used 90653-parameter models for materials A--C and 16449-parameter transfer-learned models for D--E, so its worst-material $p95$ (15.90, material D) was obtained with the 16449-parameter model. Empirical models are analytic per-material fits, whose coefficient count is not comparable to a network parameter budget (``---'').\par}
\end{table*}

\subsection{Ablation Study}
\Cref{tab:internal-main} isolates the contribution of each design choice under a shared data pipeline and evaluation script, with rows marked AWC further sharing the optimization protocol of \Cref{sec:training}, moving from the HARDCORE backbone toward CAHR-Net one axis at a time. Three axes are examined. \emph{Optimization protocol.} Applying the matched large-batch protocol (AWC) to the unchanged backbone lowers the average $p95$ from $7.47\%$ to $7.27\%$ and the worst-material $p95$ from $16.40\%$ to $15.64\%$, so a better-conditioned training region already helps before any architectural change. \emph{Condition-injection module.} Compared under the same protocol, replacing the injection with SE or a plain FiLM branch reaches $7.14\%$ and $7.85\%$ average $p95$, whereas the condition-adaptive injection of CAHR-Net attains $6.89\%$ together with the lowest average $p99$ of $11.82\%$ and the lowest worst-material $p95$ of $14.87\%$; the gap to FiLM-TCN trained at the original learning rate is consistent with the finding of \Cref{sec:training} that the modulation pathway requires the matched optimization region to become fully effective. \emph{Capacity.} Widening the network through CAHR-W12, SE+CAHR-W12, and SE+CAHR-W16 raises the parameter count from $1874$ to as many as $3057$ yet degrades the average $p95$ to the $7.69\%$--$8.47\%$ range and pushes the worst-material $p95$ above $20\%$, indicating that the improvement stems from how condition information is injected and trained rather than from added capacity. CAHR-Net therefore gives the best accuracy of the chain at a near-minimal parameter budget.

\begin{table*}[!tbp]
\caption{Ablation Study Under the Unified Experimental Chain.}
\label{tab:internal-main}
\centering
\small
\setlength{\tabcolsep}{3pt}
\renewcommand{\arraystretch}{1.15}
\resizebox{0.98\textwidth}{!}{%
\begin{tabular}{p{3.4cm} c c c c c c}
\toprule
Method & Parameters & Avg. Rel. Err. (\%) & Avg. $p95$ (\%) & Avg. $p99$ (\%) & Worst $p95$ (\%) & Rel. $\Delta p95$ vs. CAHR-Net (\%) \\
\midrule
HARDCORE & 1742 & 2.63 & 7.47 & 12.53 & 16.40 & $+8.4$ \\
HARDCORE + AWC & 1742 & 2.62 & 7.27 & 11.95 & 15.64 & $+5.5$ \\
SE-TCN + AWC + $2\times10^{-3}$ & 1875 & 2.48 & 7.14 & 11.83 & 16.23 & $+3.6$ \\
FiLM-TCN + AWC + $10^{-3}$ & 1874 & 2.70 & 7.85 & 13.19 & 17.61 & $+13.9$ \\
\textbf{CAHR-Net} & \textbf{1874} & \textbf{2.42} & \textbf{6.89} & \textbf{11.82} & \textbf{14.87} & \textbf{0.0} \\
CAHR-W12 & 2346 & 2.50 & 7.74 & 13.24 & 20.68 & $+12.3$ \\
SE + CAHR-W12 & 2524 & 2.46 & 7.69 & 12.94 & 20.04 & $+11.6$ \\
SE + CAHR-W16 & 3057 & 2.70 & 8.47 & 15.86 & 24.41 & $+22.9$ \\
\bottomrule
\end{tabular}}
\end{table*}

\subsection{Material-Level Tail Error}
\begin{figure}[!tb]
\centering
\includegraphics[width=\linewidth]{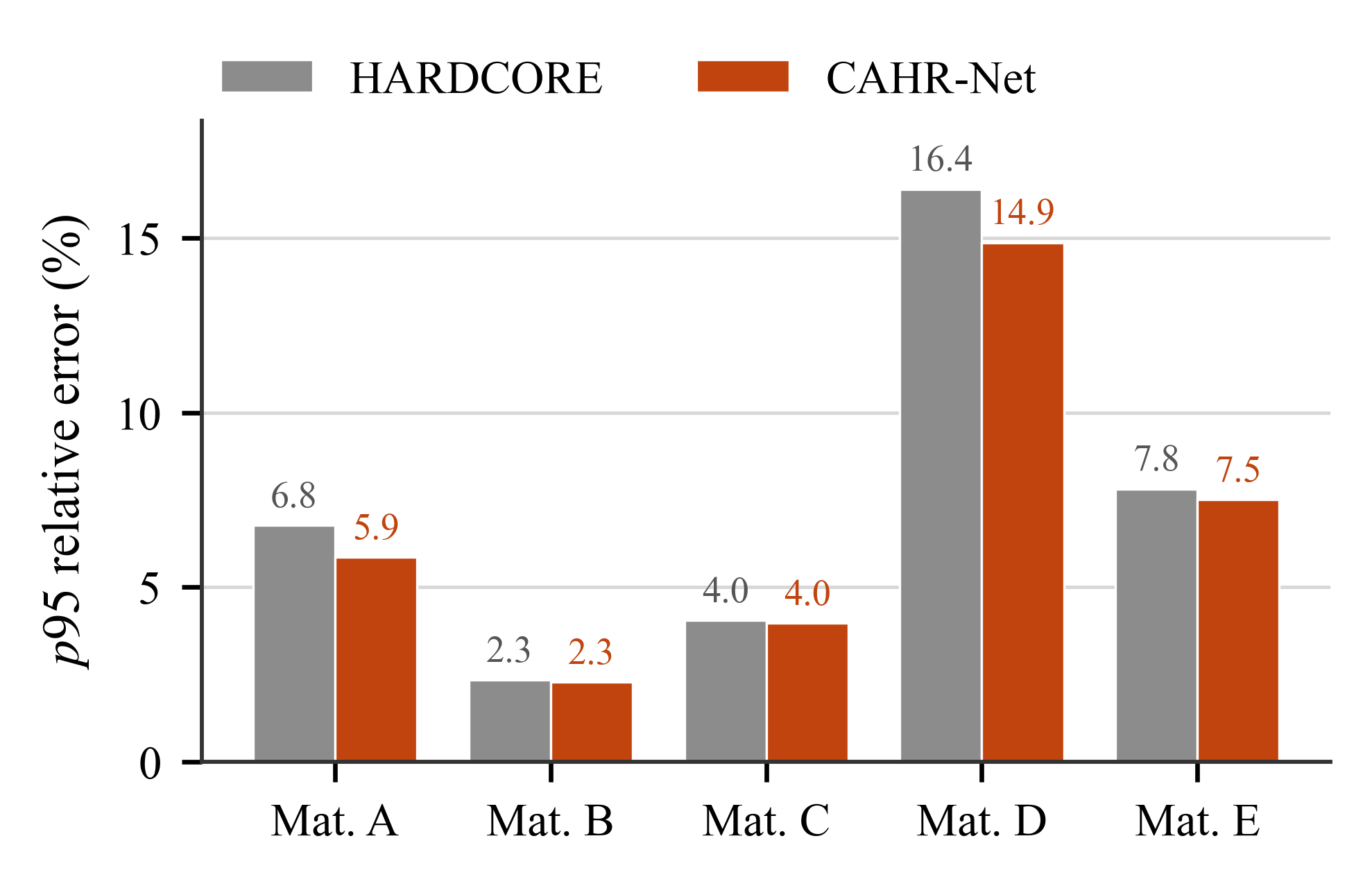}\\[4pt]
\includegraphics[width=\linewidth]{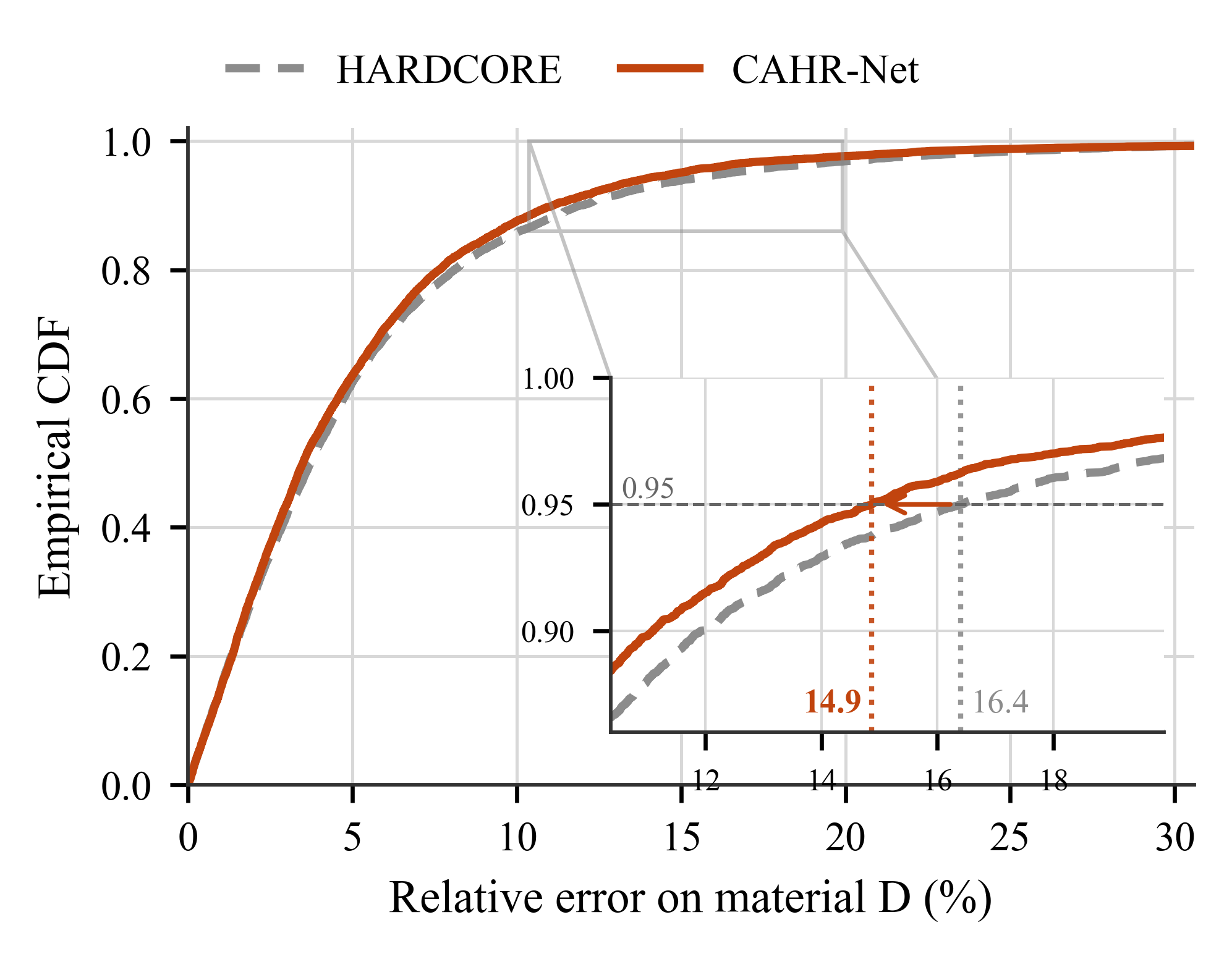}
\caption{Material-level tail analysis. (a) Material-level $p95$ comparison between HARDCORE and CAHR-Net. (b) Empirical cumulative distribution of relative error on material~D; the inset magnifies the tail region around the 95th percentile, where CAHR-Net shifts the $p95$ from $16.40\%$ to $14.87\%$.}
\label{fig:chapter6-cahr-material}
\end{figure}

\Cref{fig:chapter6-cahr-material}(a) shows that the main gain is concentrated on material~D, the most difficult material. CAHR-Net lowers the material-D $p95$ from $16.40\%$ to $14.87\%$, while materials A, B, C, and E do not show obvious degradation. This behavior is desirable because it improves the high-risk tail without transferring error to easier materials.

\Cref{fig:chapter6-cahr-material}(b) further verifies that the error distribution on material D shifts left and contracts in the high-error region. Therefore, the proposed method does not merely reduce the mean error; it directly compresses samples that would otherwise dominate design risk.

\subsection{Condition-Slice Analysis}
\begin{figure}[!tb]
\centering
\includegraphics[width=0.9\linewidth]{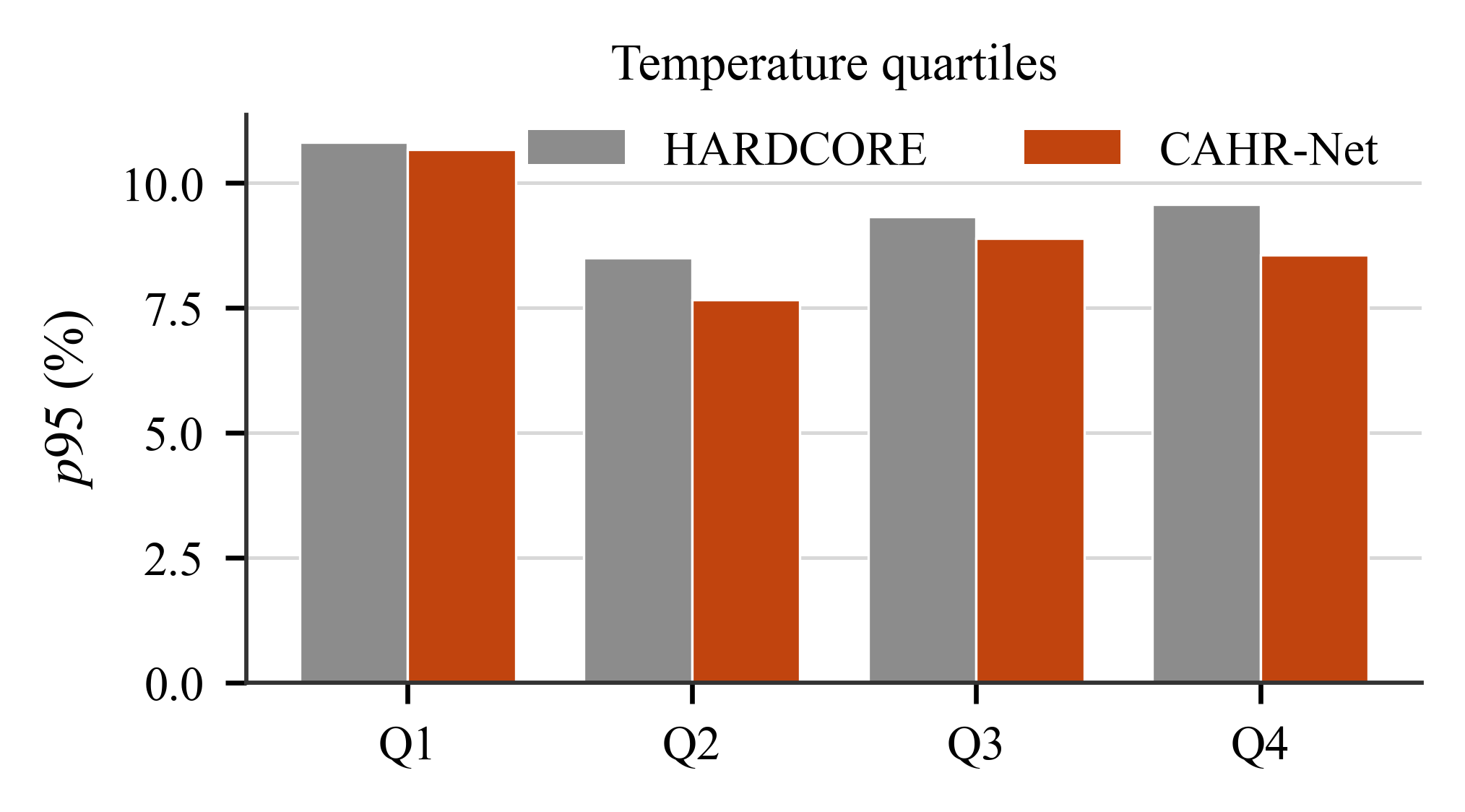}\\[3pt]
\includegraphics[width=0.9\linewidth]{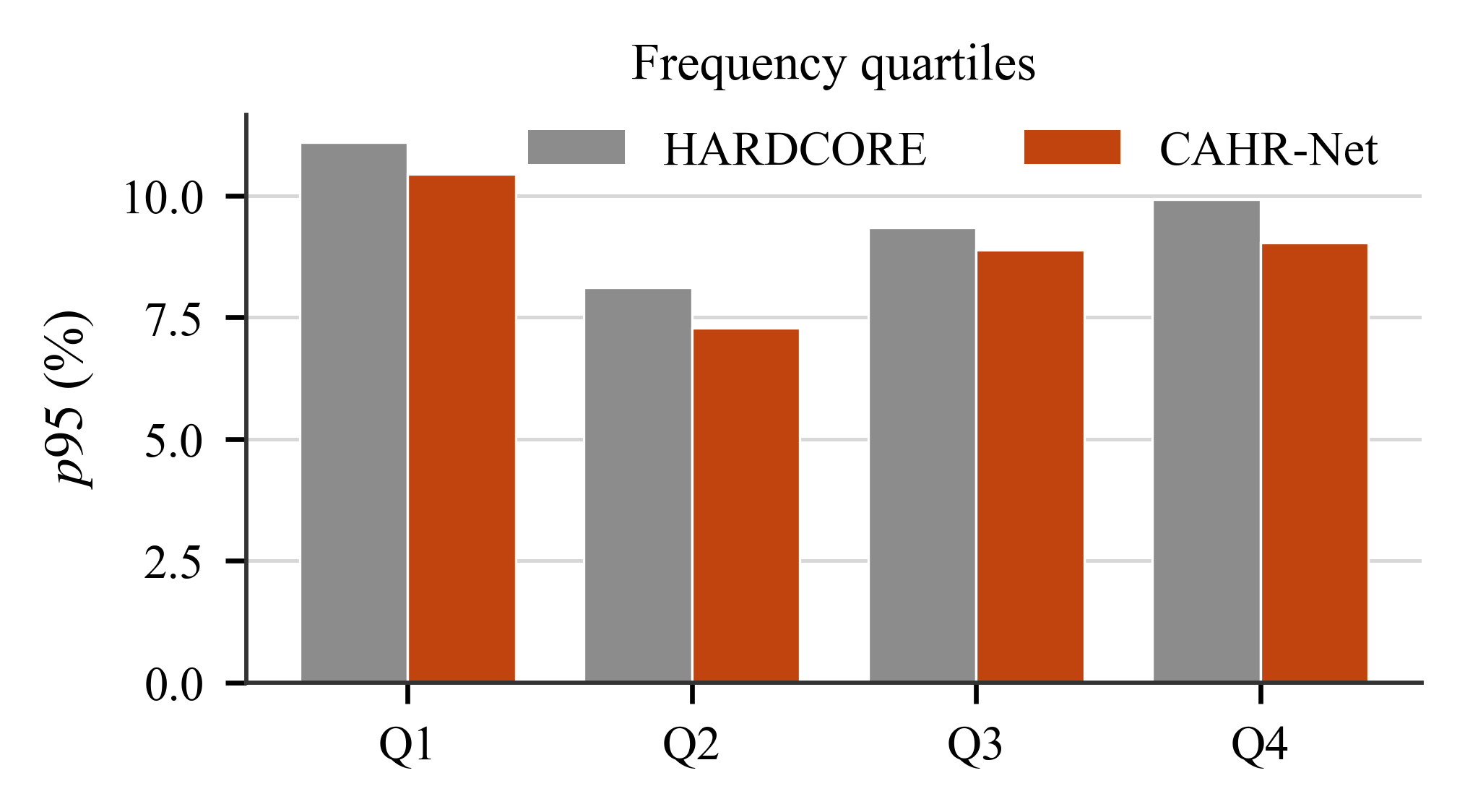}\\[3pt]
\includegraphics[width=0.9\linewidth]{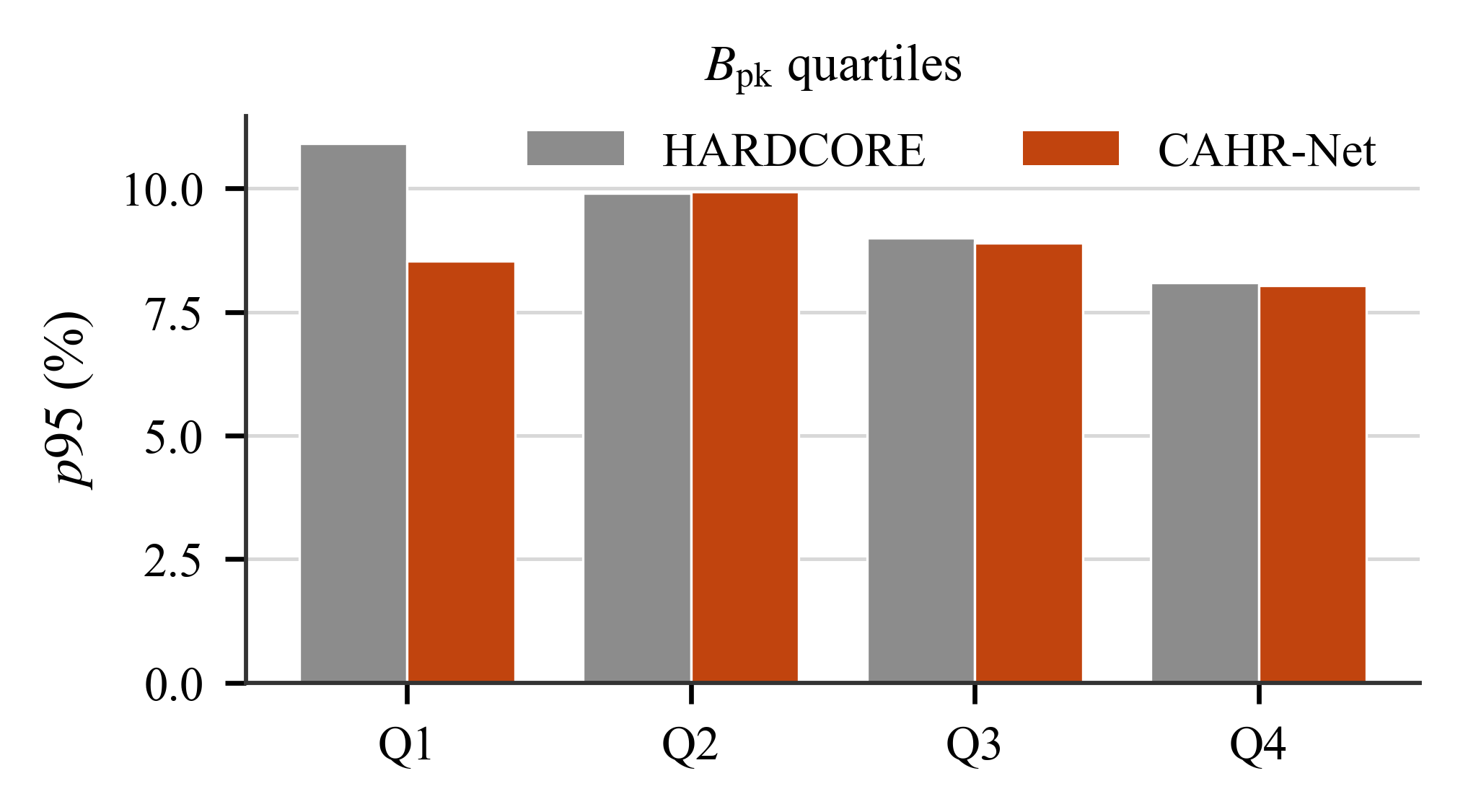}\\[3pt]
\includegraphics[width=0.9\linewidth]{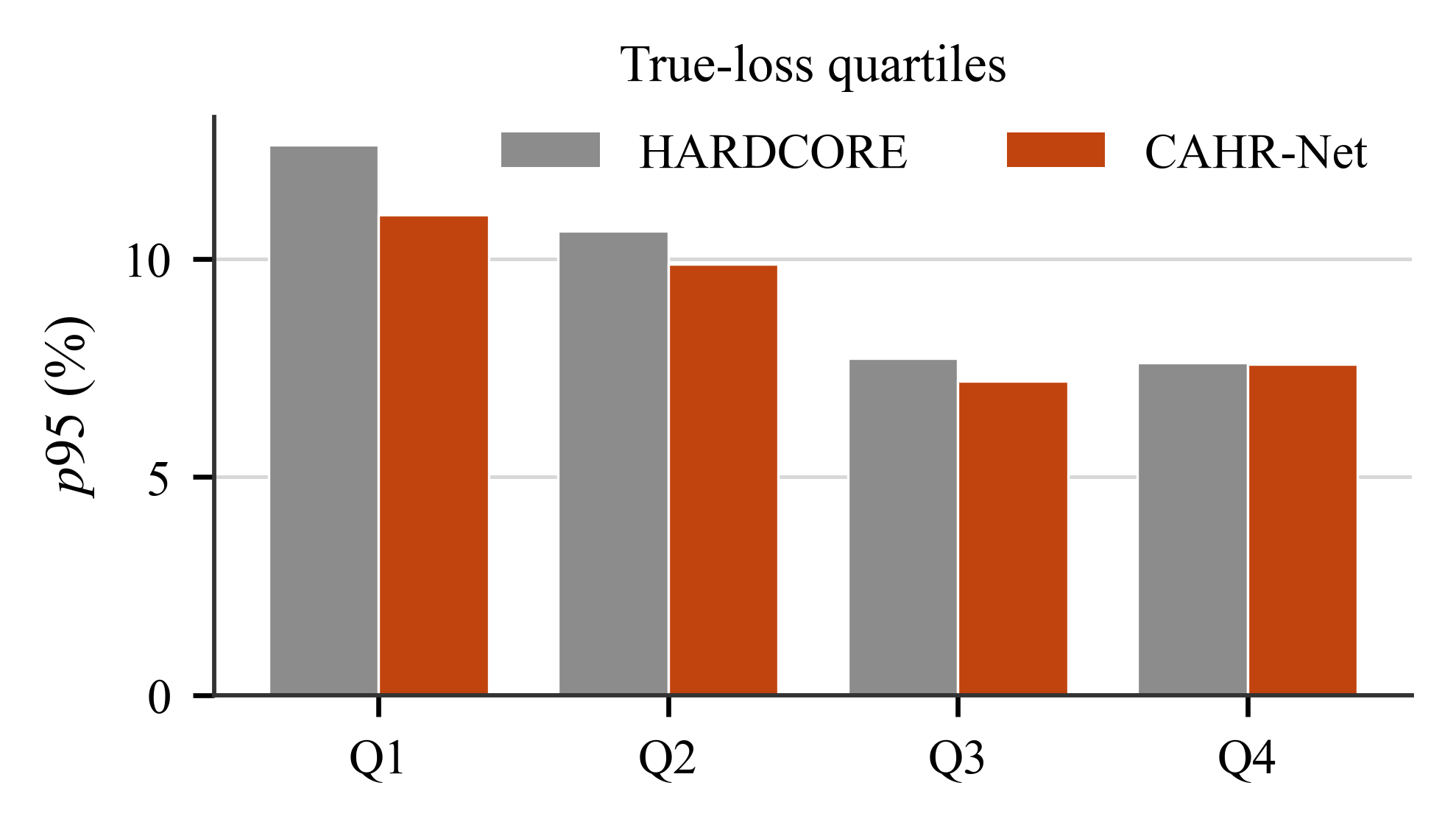}
\caption{Condition-slice analysis of $p95$ relative error, shown from top to bottom for temperature, frequency, peak flux density $B_{\mathrm{pk}}$, and true loss magnitude quartiles.}
\label{fig:chapter6-cahr-slices}
\end{figure}

To locate where the improvement appears, sample errors are pooled across the five materials and stratified by temperature, frequency, peak flux density, and true loss magnitude under shared quartile boundaries. \Cref{fig:chapter6-cahr-slices} summarizes these four condition-slice views. The gains are broad but modest: CAHR-Net lowers the $p95$ in most slices by a few tenths to about two percentage points, and holds the temperature-slice $p95$ within $7.7\%$--$10.7\%$ versus $8.5\%$--$10.8\%$ for the backbone. The clearest improvements fall on the lowest peak-flux-density quartile, from $10.91\%$ to $8.53\%$, and on the highest-error, lowest-loss-magnitude quartile, from $12.62\%$ to $11.00\%$, where relative error is intrinsically harder to control; a small number of slices, such as the second flux-density quartile, are essentially unchanged.

\subsection{Injection--Optimization Interaction and Parameter Efficiency}
\begin{figure}[!t]
\centering
\includegraphics[width=0.9\linewidth]{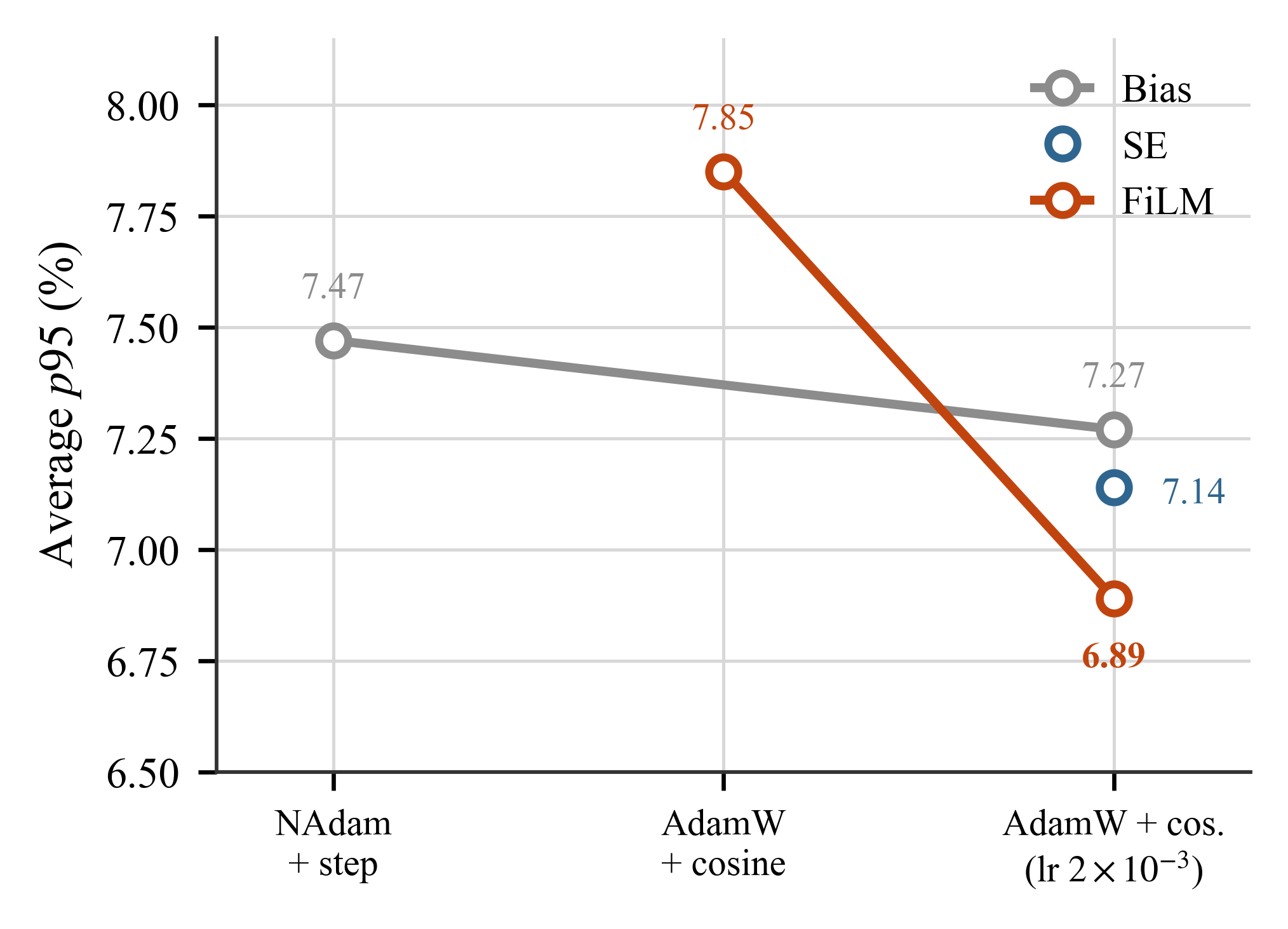}\\[3pt]
\includegraphics[width=0.9\linewidth]{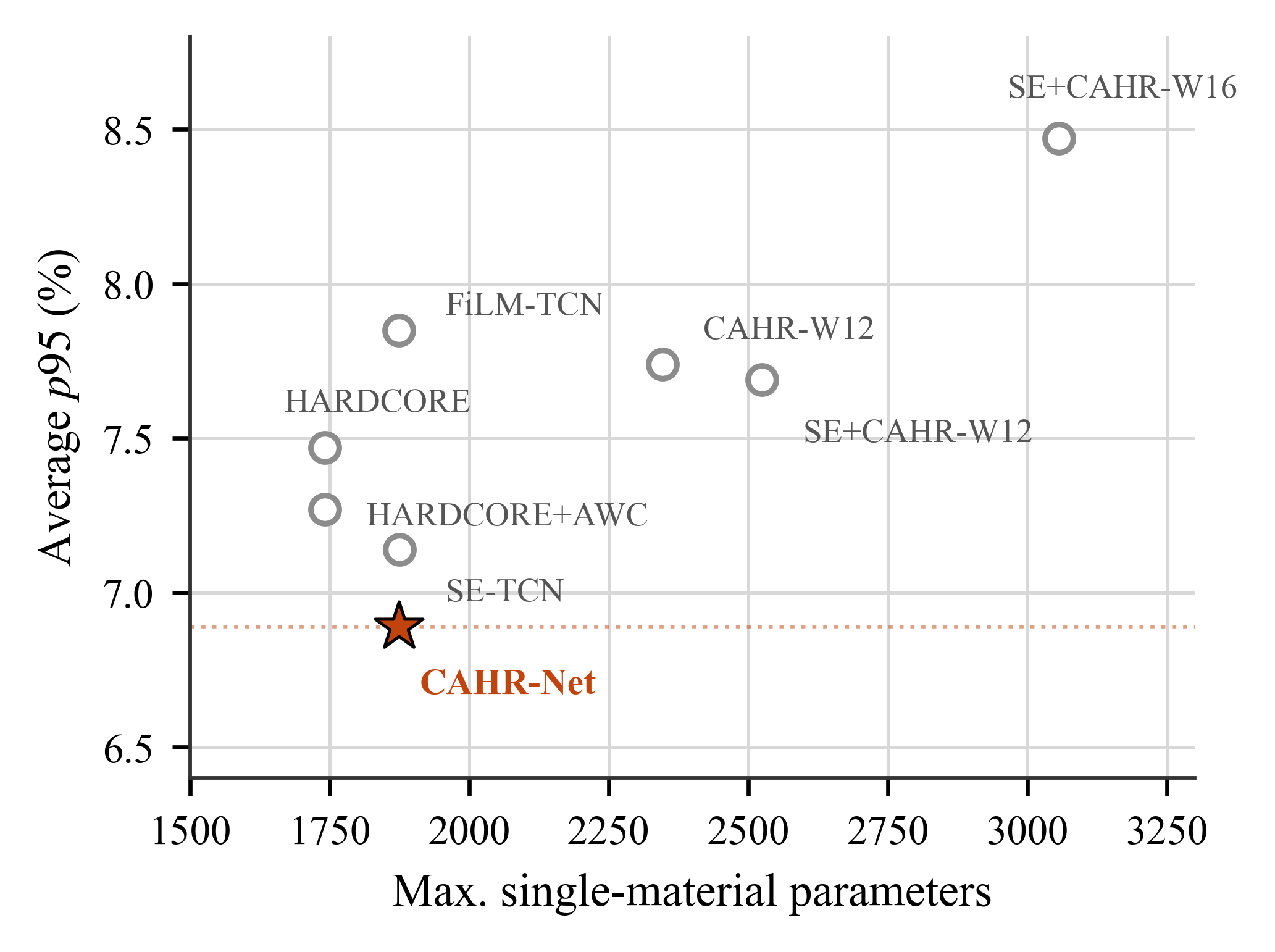}
\caption{Injection--optimization interaction and parameter efficiency. (a) Interaction between condition injection strategy and optimization protocol. (b) Parameter-count and average-$p95$ Pareto distribution of internal model variants.}
\label{fig:chapter6-cahr-ablation-pareto}
\end{figure}

\Cref{fig:chapter6-cahr-ablation-pareto}(a) presents a two-dimensional sweep over condition injection and optimization strategy. Under NAdam with step decay, bias, SE, and FiLM injection are close. After switching to AdamW with cosine scheduling, FiLM starts to show a stable advantage. Increasing the learning rate to $2\times 10^{-3}$ moves FiLM into the best region, corresponding to the CAHR-Net configuration reported in \Cref{tab:internal-main}.

\Cref{fig:chapter6-cahr-ablation-pareto}(b) shows that widening the model or adding SE-style recalibration does not reliably outperform CAHR-Net under a similar parameter budget. The dominant improvement therefore comes from how condition information enters the reconstruction process and how this structure is trained, rather than from a simple increase in capacity.

\subsection{Inference Latency Optimization}
\label{sec:latency-eval}
We now quantify the latency benefit of the accuracy-neutral deployment optimization of \Cref{sec:deployment}. \Cref{tab:onnx-latency} reports the per-sample latency of four execution backends on a single thread of a server-class Xeon Platinum 8360Y CPU, using the material-A test set and the trained checkpoint; each entry is the median over repeated timed runs, with 300 repetitions at batch size~1. Three observations follow. First, at batch size~1---the regime that matters for per-operating-point queries---switching from TorchScript to ONNX Runtime reduces the latency from $0.82$\,ms to $0.19$\,ms per sample, a $4.4\times$ speedup, and $5.6\times$ over eager execution, at zero accuracy cost. Second, the table itself diagnoses where the gain comes from: at batch size~256 all four backends converge to approximately $150\,\mu$s per sample, which is the arithmetic floor of the model on this core; consequently, about $82\%$ of the batch-1 TorchScript time is per-invocation framework overhead rather than computation, and the runtime switch removes precisely this overhead instead of approximating the model. Third, at batch size~2048 the PyTorch-based backends degrade to roughly $370\,\mu$s per sample whereas ONNX Runtime remains at $174\,\mu$s, indicating better operator fusion and memory locality at large batches.

To assess whether these gains are tied to the proposed architecture, the identical export-and-verification procedure is applied to the bias-injection backbone HARDCORE, and \Cref{tab:model-latency} contrasts the two models before and after the backend switch. The backbone exhibits the same $4.4\times$ reduction in single-sample latency, from $798$ to $182\,\mu$s, which indicates that the optimization exploits a property shared by this class of compact loop-reconstruction models---their inference time is dominated by framework dispatch rather than by computation---and hence generalizes beyond CAHR-Net itself. \Cref{tab:model-latency} also shows that the latency of the two models remains within $5\%$ of each other at every batch size under the optimized backend---$186$ versus $182\,\mu$s at batch size~1---an absolute difference of a few microseconds per sample. Condition-adaptive modulation thus incurs no practically relevant serving cost: the accuracy improvements of \Cref{tab:internal-main} are obtained at essentially the latency of the unmodulated backbone.

\begin{table}[!tb]
\caption{Single-Thread CPU Inference Latency of CAHR-Net Across Execution Backends ($\mu$s per Sample, Material-A Test Set)}
\label{tab:onnx-latency}
\centering
\small
\setlength{\tabcolsep}{3pt}
\renewcommand{\arraystretch}{1.15}
\resizebox{\linewidth}{!}{%
\begin{tabular}{c c c c c}
\toprule
Batch Size & PyTorch Eager & TorchScript & TorchScript (Opt.) & ONNX Runtime \\
\midrule
1 & 1051 & 822 & 788 & \textbf{186} \\
16 & 200 & 185 & 169 & \textbf{149} \\
256 & 161 & 154 & \textbf{152} & \textbf{152} \\
2048 & 366 & 372 & 372 & \textbf{174} \\
\bottomrule
\end{tabular}}
\end{table}

\begin{table}[!tb]
\caption{Model-Level Latency Before and After Backend Optimization ($\mu$s per Sample, Single CPU Thread)}
\label{tab:model-latency}
\centering
\small
\setlength{\tabcolsep}{3pt}
\renewcommand{\arraystretch}{1.15}
\resizebox{\linewidth}{!}{%
\begin{tabular}{c c c c}
\toprule
Batch Size & HARDCORE (TorchScript) & CAHR-Net Before Opt.\ (TorchScript) & CAHR-Net After Opt.\ (ONNX Runtime) \\
\midrule
1 & 798 & 822 & 186 \\
16 & 167 & 185 & 149 \\
256 & 147 & 154 & 152 \\
2048 & 354 & 372 & 174 \\
\bottomrule
\end{tabular}}
\end{table}

From a deployment perspective, $0.19$\,ms per sample on one CPU thread corresponds to more than $5000$ loss evaluations per second per core, which is sufficient to embed the model directly in interactive design iteration or in the inner loop of a circuit-level optimization sweep over thousands of operating points, with no GPU in the serving path. The exported ONNX graph, together with the roughly $7.5$\,KB of FP32 weights, is also the standard entry format for embedded inference toolchains, so it constitutes a ready artifact for microcontroller-class deployment. Two caveats bound this result: the reported latencies are measured on a server-class CPU core rather than on a microcontroller, and further reductions through integer quantization or sequence downsampling are left as future work.

\section{Conclusion}
\label{sec:conclusion}
This paper returns to a fundamental question: where, in a magnetic core loss model, should operating conditions enter? CAHR-Net answers by placing them in the hysteresis-loop reconstruction process rather than at the model output. The network predicts along the physical $B(t)\rightarrow \hat{H}(t)\rightarrow BH\rightarrow \Phat$ chain, keeping the hysteresis loop as an explicit modeling object throughout. On this basis, FiLM converts frequency, temperature, and waveform statistics into channel-wise scales and shifts that act directly on the intermediate representation from which the magnetic-field waveform is reconstructed---the point at which these conditions physically act---instead of merely correcting a scalar output at the end of the model. The training protocol is presented together with the network architecture because the two are inseparable: under the original NAdam and step-decay configuration, the modulation pathway remains in a near-identity regime; only under the matched protocol combining AdamW, cosine scheduling, and a larger learning rate does it leave that regime and become effective. CAHR-Net therefore derives its value not from any single dimension, but from the coupling of physical loop reconstruction, structured condition modulation, and matched optimization: within a model budget on the order of $10^3$ parameters, it retains intermediate quantities that engineers can compare directly with measured hysteresis loops while achieving the tail accuracy attainable by black-box models.

Under the MagNet A--E final protocol, CAHR-Net attains the lowest average $p95$ among the compared methods---$6.89\%$ with 1874 parameters, versus $7.78\%$ for the strongest black-box entry at about $48\times$ the parameter count, which it also surpasses on worst-material $p95$---and offers an operating point that neither black-box nor existing gray-box models cover: a material-D $p95$ compressed from $16.40\%$ to $14.87\%$ relative to the reconstruction backbone, and a prediction chain whose intermediate quantities remain electromagnetically readable. Ablations attribute this gain to the synergy of loop reconstruction, structured condition modulation, and matched optimization. The accuracy--interpretability combination is also cheap to serve: after an accuracy-neutral export to a dedicated inference runtime, single-sample CPU latency drops from $0.82$\,ms to $0.19$\,ms on one thread, so the model can run inside design loops without an accelerator.

This study remains limited to per-material training on the five MagNet ferrites and single-period steady-state excitation without DC bias. Future work will address these issues together with material recommendation, uncertainty-aware design margins, and deployment-oriented model compression.

\bibliographystyle{IEEEtran}
\bibliography{reference}

\end{document}